\documentclass[final]{article}
\usepackage{neurips_2025}
\usepackage[utf8]{inputenc} 
\usepackage[T1]{fontenc}    
\usepackage{hyperref}       
\usepackage{url}            
\usepackage{booktabs}       
\usepackage{amsfonts}       
\usepackage{nicefrac}       
\usepackage{microtype}      
\usepackage{xcolor} 
\usepackage{algorithm}
\usepackage{algorithmic}
\usepackage{amsmath}
\usepackage{graphicx}
\usepackage{amssymb}
\usepackage{subcaption}
\usepackage{booktabs}
\usepackage{natbib} 
\usepackage{bm} 
\usepackage{tikz}
\usepackage{longtable}
\usepackage{array}
\usetikzlibrary{shapes.geometric, arrows, positioning}
\usepackage{indentfirst}

\usepackage{enumitem}

\title{Learning to Route: Per-Sample Adaptive Routing for Multimodal Multitask Prediction}

\author{%
\textbf{Marzieh Ajirak$^1$} \quad
\textbf{Oded Bein$^1$} \quad
\textbf{Ellen Rose Bowen$^1$} \quad
\textbf{Dora Kanellopoulos$^1$} \\
\textbf{Avital Falk$^1$} \quad
\textbf{Faith Gunning$^1$} \quad
\textbf{Nili Solomonov$^1$}\thanks{Co–senior authors.} \quad
\textbf{Logan Grosenick$^{1,2}$}\footnotemark[1] \\
$^1$Department of Psychiatry, Weill Cornell Medicine\\
$^2$Feil Family Brain \& Mind Research Institute, Weill Cornell Medicine
}

\begin{document}
\maketitle

\begin{abstract}
We propose a unified framework for adaptive routing in multitask, multimodal prediction settings where data heterogeneity and task interactions vary across samples. Motivated by applications in psychotherapy where structured assessments and unstructured clinician notes coexist with partially missing data and correlated outcomes, we introduce a routing-based architecture that dynamically selects modality processing pathways and task-sharing strategies on a per-sample basis. Our model defines multiple modality paths including raw and fused representations of text and numeric features and learns to route each input through the most informative expert combination. Task-specific predictions are produced by shared or independent heads depending on the routing decision, and the entire system is trained end-to-end. We evaluate the model on both synthetic data and real-world psychotherapy notes predicting depression and anxiety outcomes. Our experiments show that our method consistently outperforms fixed multitask or single-task baselines, and that the learned routing policy provides interpretable insights into modality relevance and task structure. This addresses critical challenges in personalized healthcare by enabling per-subject adaptive information processing that accounts for data heterogeneity and task correlations. Applied to psychotherapy, this framework could improve mental health outcomes, enhance treatment assignment precision, and increase clinical cost-effectiveness through personalized intervention strategies.
\end{abstract}

\section{Introduction}
Modern predictive models increasingly operate in settings with multiple heterogeneous input modalities and correlated outputs. In domains such as clinical informatics and behavioral health, data arrive in diverse formats that include structured numerical features (clinical scales, sensor measurements) and unstructured text (clinician notes, patient narratives) \citep{baltruvsaitis2018multimodal, rajkomar2018scalable}. These modalities differ in structure, coverage, semantic density, and sample-level informativeness. Predictive targets, for example, depression and anxiety scores, often correlate but do not overlap entirely. This combination motivates multimodal multitask learning models that integrate heterogeneous inputs while capturing structured relationships across tasks.

Multimodal learning aims to learn joint representations from diverse data sources. However, most existing approaches assume fixed fusion strategies and complete modality availability \citep{ruder2017overview,liu2022multimodal}. Similarly, multitask learning (MTL) typically relies on globally shared architectures, applying the same parameter sharing scheme to all inputs. However, these assumptions are often violated in real-world settings. Modality quality and informativeness can vary substantially across samples, and task relationships may differ depending on latent factors such as individual behavior, context, or data completeness. Ignoring these forms of heterogeneity leads to suboptimal representations, and reduced predictive performance and generalization.

To address these limitations, we propose a unified framework that performs adaptive expert routing over both modality and task configurations \citep{shazeer2017outrageously, ma2018modeling, rosenbaum2017routing}. Our model defines a set of modality transformation paths including unimodal and fused representations and a set of prediction heads corresponding to single-task and multitask supervision structures. A learnable routing mechanism selects a personalized expert pathway for each input by modeling a probabilistic mixture over modality-task combinations, capturing both input-dependent modality preferences and task-level interaction structure. This approach generalizes conditional computation and mixture-of-experts (MoE) frameworks to settings where latent structure exists across inputs, supervision targets, and representational hierarchies.

The routing policy is parameterized by neural networks and trained jointly with all expert modules using backpropagation. To promote policy diversity and mitigate expert collapse (where a subset of experts get nearly all the traffic), we incorporate an entropy regularization term \citep{fedus2022switch}. This design enables sample-specific selection of both data representations and task decoders, effectively adapting the computation graph to underlying data geometry and supervision structure.

We validate the proposed framework through a series of experiments within synthetic and real world data. In synthetic data with controlled variation in modality relevance and task correlation, our model outperforms fixed multitask and single-task baselines while recovering interpretable routing policies. In real-world psychotherapy data with structured assessments and unstructured clinician notes \citep{benton2017multitask, niu2024ehr}, the model improves prediction of anxiety and depression outcomes and reveals routing decisions that align with intuitive task-modality interactions. These results demonstrate that adaptive routing over representation and supervision structure is a powerful mechanism for modeling heterogeneous, multimodal prediction tasks.

In summary, our \textbf{major contributions} are: (1) we develop a modular architecture supporting multiple modality transformation paths and adaptive task-sharing schemes, (2) we design a probabilistic routing mechanism that dynamically selects, for each input, both optimal modality pathways and task configurations based on input and output structure, and (3) we demonstrate significant improvements in both prediction accuracy and interpretability across synthetic and real-world clinical datasets. Our approach consistently outperforms standard multitask and single-task baselines, with immediate potential applications for enhancing decision support in mental healthcare and broader medical contexts. The framework's ability to adapt to heterogeneous data while modeling structured relationships across tasks makes it particularly valuable for real-world clinical environments where data heterogeneity and quality varies (across patients, clinicians, sites, etc.) and outcome measures are often interdependent (e.g. multiple clinical scales or physiological measurements).


\section{Related work}
\subsection{Multimodal learning in clinical and mental health contexts}

A growing body of research investigates multimodal learning with structured clinical data (e.g., electronic health records, standardized assessments) and unstructured text (e.g., clinician notes) to improve outcome prediction. In general medical AI, combining tabular EHR features with narrative notes has led to measurable gains in predictive performance \citep{lyu2022multimodal}. In the mental health domain, fusion of structured and unstructured data has yielded similar benefits. For instance, \citet{garriga2023combining} predicted 28-day psychiatric crisis risk using both structured EHR variables and clinical note text, reporting that models leveraging both modalities outperformed unimodal baselines. Other studies have found that incorporating text embeddings derived from models like BERT into structured-input pipelines improves accuracy across various clinical tasks \citep{ye2024multimodal}.

However, the gains from multimodal fusion are not consistent across settings. For example, \citet{kotula2025comparison} found that augmenting vital signs and lab values with concept-extracted notes led to only marginal improvements for ICU deterioration prediction. These mixed results suggest that the informativeness of unstructured text and the chosen fusion strategy can critically impact performance. Our approach addresses this limitation by supporting adaptive fusion. Instead of using a fixed integration schema, the model determines how to combine structured and unstructured inputs separately for each sample. This enables more flexible use of the available modalities, depending on their informativeness for the individual case.

\subsection{Multitask learning for mental health prediction}
Multitask learning (MTL) offers a principled approach for modeling multiple correlated clinical outcomes, particularly in mental health prediction tasks \citep{ma2018modeling}. Conditions such as major depressive disorder and generalized anxiety frequently co-occur and exhibit overlapping symptom profiles, yet clinical models often treat them as independent targets or apply joint statistical models that fail to capture their structured dependencies. MTL enables parameter sharing across tasks, which supports the learning of shared representations that can improve generalization and predictive performance. Prior work has demonstrated the effectiveness of this approach in mental health settings. For example, \citet{buddhitha2023multi} constructed a shared encoder with both hard and soft parameter sharing to jointly model mental illness and suicide ideation risk, outperforming task-specific baselines. In another study, \citet{saylam2024multitask} showed that jointly predicting depression, anxiety, and stress levels  led to performance gains for depression and stress compared to single-task models. These findings show that MTL architectures with shared encoders and task-specific decoders can leverage cross-task structure while preserving task specialization.

Task relatedness plays a central role in the effectiveness of multitask learning. When tasks are closely correlated or reflect underlying causal relationships, multitask models often achieve positive transfer. In contrast, unrelated tasks can lead to negative transfer and representation interference \citep{ma2018modeling}. To address this, recent work has introduced methods that learn task relationships directly from data. The Multi-gate Mixture-of-Experts (MMoE) architecture, for example, assigns each task a dedicated gating network that selects among a shared pool of experts, thereby enabling flexible combinations of shared and task-specific computation \citep{ma2018modeling}. These designs relax the binary distinction between fully shared and fully independent decoders and support a continuum of sharing that adapts to task similarity. In the context of mental health prediction, where tasks often reflect partially overlapping symptom dimensions, such flexibility is especially important. We build on this idea by introducing an adaptive routing mechanism that integrates with MTL and selects the appropriate degree of sharing across tasks based on sample-level signals.

\subsection{Adaptive routing and mixture-of-experts for heterogeneous inputs/tasks}
Recent advances in adaptive routing \citep{rosenbaum2017routing} and Mixture-of-Experts (MoE) \citep{jacobs1991adaptive} architectures offer flexible strategies for modeling heterogeneity across both inputs and tasks. MoE models consist of multiple specialized sub-networks (“experts”)  and a trainable router that assigns each input to a subset of these experts based on its characteristics \citep{mu2025comprehensive,yuhao2024gaussian}. This setup enables conditional computation, where different experts can focus on different regions of the input space or specialize in particular tasks. In large-scale language and vision models, sparsely activated MoEs have improved computational efficiency and generalization compared to densely connected alternatives \citep{zhou2022mixture}.

In clinical applications, conditional routing offers practical benefits for handling input variability. For instance, a model can assign a sample with detailed textual notes to a text-specialized expert, while directing a sample with only numerical features to a different expert. This flexibility supports personalized prediction pipelines without requiring separate models for every data configuration. Recent work in medical machine learning has applied these principles to multimodal tasks. The dynamic routing framework proposed by \citet{wu2025dynamic} selects modality-task combinations on a per-sample basis, capturing the dependencies between clinical outcomes and input modalities. Their model learns a modality fusion strategy using mutual information regularization, which guides the decomposition of each sample’s data into shared and distinct components.

Routing mechanisms have also been applied to task-level adaptation. For example, \citet{rosenbaum2017routing} introduced routing networks that learn input-dependent paths through modular function blocks. This architecture allows the model to activate shared components when beneficial and fall back on task-specific routes when task interference arises. These methods collectively demonstrate the value of flexible routing schemes in domains with complex, variable inputs and overlapping prediction objectives. Our work builds on this foundation by integrating sample-level routing over both modalities and tasks, enabling a unified framework for adaptive multimodal multitask learning.

\section{Methodology}
\subsection{Bidirectional Transformation Between Structured and Unstructured Modalities}
\label{sec:modality}
Our approach addresses the multimodal nature of psychotherapy datasets by introducing flexible, bidirectional transformations between structured numerical data and unstructured text. Rather than treating each modality as isolated, we design a unified architecture capable of transforming and integrating representations across formats. This enables the model to operate uniformly across patients with different modality availability, data missingness, or data quality.

Let $X_{\text{num}} \in \mathbb{R}^{d_{\text{num}}}$ represent the structured numerical input, and $X_{\text{text}} \in \mathcal{T}$ denote unstructured textual data. We define two learned transformation functions:
\begin{align}
f_{\text{num2text}} &: \mathbb{R}^{d_{\text{num}}} \rightarrow \mathcal{T}, \\
f_{\text{text2num}} &: \mathcal{T} \rightarrow \mathbb{R}^{d_{\text{text}}},
\end{align}
where $f_{\text{num2text}}$ converts numerical input into semantically meaningful natural language, and $f_{\text{text2num}}$ encodes text into numerical embeddings. Here, $f_{\text{text2num}}$ can be implemented using sentence embedding models such as MPNet \citep{song2020mpnet} or Sentence-BERT \citep{reimers2019sentence}. The generated embedding is a structured numerical representation that can be integrated with traditional statistical models or deep learning architectures designed for numerical inputs.

We define the following four cases:

\begin{enumerate}
    \item \textbf{Text-only (T1):} Use only $X_{\text{text}}$. Numerical input is ignored or unavailable.
    
    \item \textbf{Numerical-only (N1):} Use only $X_{\text{num}}$. Text input is ignored or unavailable.
    
    \item \textbf{Textualized Numerical + Text (T2):} Apply $f_{\text{num2text}}(X_{\text{num}})$ to produce $X^{(\text{text})}_{\text{num}}$, and concatenate with original text:
    \begin{equation}
        X = \text{concat}_{\text{text}}(X^{(\text{text})}_{\text{num}}, X_{\text{text}}),
    \end{equation}
    forming a unified textual input passed to a text-native model (e.g., transformer). This allows us to employ pretrained language models such as BERT \citep{devlin2018bert} and MPNet \citep{song2020mpnet} to process the text data.
    
    \item \textbf{Numerical + Text Embedding (N2):} Apply $f_{\text{text2num}}(X_{\text{text}})$ to produce $X^{(\text{num})}_{\text{text}}$, and concatenate with original numerical features:
    \begin{equation}
        X = \text{concat}_{\text{num}}(X^{(\text{num})}_{\text{text}}, X_{\text{num}}),
    \end{equation}
    resulting in a unified numerical representation passed to a numerical backbone.
\end{enumerate}

These paradigms enable flexible fusion strategies that are adaptable to data quality, availability, and downstream model compatibility. Figure \ref{fig:translate} illustrates these conversions and concatenation.  The two-way conversion creates a flexible bridge between data types. It lets structured data be understood in natural language terms (textualization) while also transforming text into numerical formats that work with traditional data models.

\subsection{Multitask vs. Single-task Learning Objectives}
We adopt a multitask learning (MTL) framework to address correlated clinical outcomes (depression, anxiety) of psychotherapy \citep{ruder2019neural}. The performance of commonly used multitask models often depends on the relationships between tasks. Therefore, studying the trade-offs between task-specific objectives and inter-task dependencies is crucial. We focus on predicting two common key clinical outcomes: Depression (measured with the Patient Health Questionnaire-9 [PHQ-9] \citep{kroenke2001phq}) and anxiety (Generalized Anxiety Disorder-7 [GAD7] \citep{spitzer2006brief}). The correlation between these measures provides distinct and unique information about a patient's symptoms. Our approach supports both single-task learning (STL) and multitask learning (MTL):

\begin{enumerate}

    \item \textbf{Single-Task Learning (STL):} Trains two independent models, one per outcome. Each model $f_k(X^{(i)})$ predicts both the outcome $\hat{y}_k$ and its log-variance $\log \sigma_k^2$, where $k \in \{\text{PHQ}, \text{GAD}\}$ and $X^{(i)}$ is the modality-transformed input for paradigm $i \in \{\text{T1}, \text{N1}, \text{T2}, \text{N2}\}$. For each task:
    \begin{equation}
        f_k(X^{(i)}) = \left( \hat{y}_k(X^{(i)}), \log \sigma_k^2(X^{(i)}) \right),
    \end{equation}
    the heteroscedastic loss for task $k$ is:
    \begin{equation}
        \mathcal{L}_{\text{STL}, k}(X^{(i)}, y_k) = 
        \frac{1}{2} \cdot \frac{(y_k - \hat{y}_k(X^{(i)}))^2}{\sigma_k^2(X^{(i)})} 
        + \frac{1}{2} \log \sigma_k^2(X^{(i)}),
    \end{equation}
    and the total STL loss is:
    \begin{equation}
        \mathcal{L}_{\text{STL}} = \mathcal{L}_{\text{STL}, \text{PHQ}} + \mathcal{L}_{\text{STL}, \text{GAD}}.
    \end{equation}

    \item \textbf{Multi-Task Learning (MTL):} Uses a shared encoder followed by task-specific heads. The shared encoder produces a latent representation $z = f_{\text{shared}}(X^{(i)})$.

    Each task head then outputs both a mean prediction and log-variance:
    \begin{align}
        f_k(z) &= \left( \hat{y}_k(z), \log \sigma_k^2(z) \right), \quad k \in \{\text{PHQ}, \text{GAD} \},
    \end{align}

    with the heteroscedastic loss per task being:
    \begin{equation}
        \mathcal{L}_{\text{MTL}, k}(z, y_k) = 
        \frac{1}{2} \cdot \frac{(y_k - \hat{y}_k(z))^2}{\sigma_k^2(z)} 
        + \frac{1}{2} \log \sigma_k^2(z),
    \end{equation}

    and the total MTL loss is:
    \begin{equation}
        \mathcal{L}_{\text{MTL}} = \sum_{k \in \{\text{PHQ}, \text{GAD}\}} \mathcal{L}_{\text{MTL}, k}(z, y_k).
    \end{equation}

\end{enumerate}

MTL improves generalization when tasks share underlying signals, while STL is preferred if task-specific features dominate or tasks conflict.


\subsection{Modeling sample heterogeneity via probabilistic expert routing}
Psychotherapy data includes inherent heterogeneity. First, there is variation across patients due to both individual heterogeneity in patient biology and symptom presentation, as well as due to data modality, missingness, and quality per patient, clinician, or site. Second, there is variability in the extent to which measured outcomes correlate. For example, some patients may present with both depression and anxiety symptoms that change in tandem during treatment, while others may experience specific changes to just anxiety or depression symptoms. Rather than applying a fixed modality paradigm or learning strategy across all samples, we assume samples vary in modality informativeness and task relevance. To model this heterogeneity, we introduce a hierarchical mixture-of-experts architecture that probabilistically routes each sample to one of eight expert paths: $\{\text{T1, T2, N1, N2}\} \times \{\text{STL, MTL}\}$.

\paragraph{Routing Architecture:}

We define a two-stage probabilistic routing mechanism that dynamically selects among modality paths and task setups on a per-sample basis.

\begin{itemize}
    \item \textbf{Modality Router:} {Given $X^{(i)}$,} a gating function $r_{\text{mod}}$ outputs a probability distribution over the four modality paths (T1, T2, N1, N2):
  \begin{equation}
        \pi_{\text{mod}} = \text{softmax}(r_{\text{mod}}(X_{\text{num}}, X_{\text{text}})) \in \mathbb{R}^4.
    \end{equation}
    Each modality path \( i \in \{1, 2, 3, 4\} \) corresponds to a specific transformation of the input, resulting in a modality-specific representation \( X^{(i)} \).
    
    \item \textbf{Task Router:} For each modality-transformed input \( X^{(i)} \), a second gating function $r_{\text{task}}$ computes a distribution over task strategies (STL or MTL):
    \begin{equation}
     { \pi_{\text{task}}^{(i,j)}} = \text{softmax}(r_{\text{task}}(X^{(i)})) \in \mathbb{R}^2.
    \end{equation}
    Here, \( j \in \{1, 2\} \) indexes STL and MTL, respectively.
\end{itemize}

Each combination of modality path \( i \) and task strategy \( j \) defines an expert model \( f^{(i,j)} \) that outputs heteroscedastic predictions for both outcomes:
\[
f^{(i,j)}(X^{(i)}) = \left(
\hat{y}^{(i,j)}_{\text{PHQ}}, \log \sigma^{2(i,j)}_{\text{PHQ}},
\hat{y}^{(i,j)}_{\text{GAD}}, \log \sigma^{2(i,j)}_{\text{GAD}}
\right).
\]

The overall predictive outputs are computed as a mixture over expert paths:
\begin{align}
\hat{y}_{\text{PHQ}} &= \sum_{i=1}^{4} \sum_{j=1}^{2} 
\pi_{\text{mod}}^{(i)} \cdot \pi_{\text{task}}^{(i,j)} \cdot \hat{y}^{(i,j)}_{\text{PHQ}}, \\
\hat{y}_{\text{GAD}} &= \sum_{i=1}^{4} \sum_{j=1}^{2} 
\pi_{\text{mod}}^{(i)} \cdot \pi_{\text{task}}^{(i,j)} \cdot \hat{y}^{(i,j)}_{\text{GAD}}.
\end{align}

Each expert also contributes to the total uncertainty-aware loss. For soft routing, we compute the expected loss across all expert paths:
\begin{equation}
\mathcal{L}_{\text{total}} = \sum_{k \in \{\text{PHQ}, \text{GAD}\}} \sum_{i=1}^{4} \sum_{j=1}^{2}
\pi_{\text{mod}}^{(i)} \cdot \pi_{\text{task}}^{(i,j)} \cdot 
\left[
\frac{1}{2} \cdot \frac{(y_k - \hat{y}_k^{(i,j)})^2}{\sigma_k^{2(i,j)}} 
+ \frac{1}{2} \log \sigma_k^{2(i,j)}
\right].
\end{equation}

In the case of hard routing, we replace the mixture with discrete selection and only one \( (i, j) \) pair contributes to the prediction and loss. We explore both hard routing (differentiable approximation to the argmax operator) and soft routing (mixture), trained end-to-end using total task loss. In soft routing, gradients flow through all paths, encouraging specialization. In hard routing, selection is treated as discrete via Gumbel-Softmax reparameterization \citep{jang2016categorical}. 
This architecture captures complex heterogeneity in both data representation and outcome structure, automatically discovering which modality and learning scheme are best suited for each sample.

\begin{figure}
    \centering
    \includegraphics[width=0.6\linewidth]{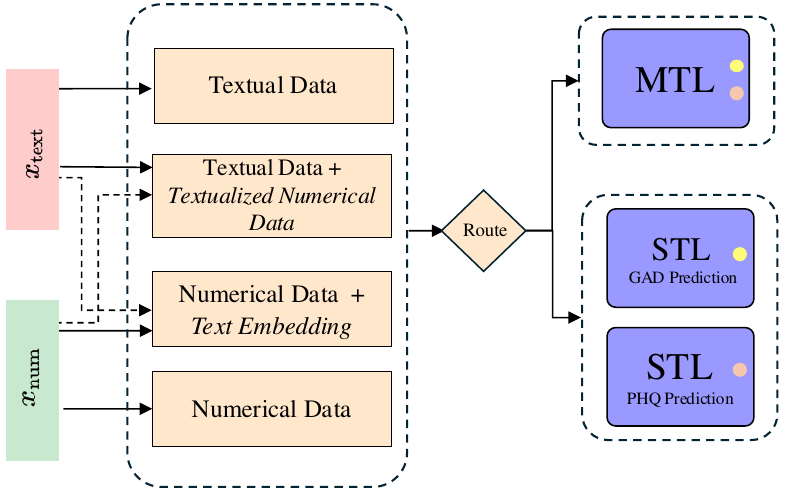}
    \caption{Overview of our adaptive machine learning framework for mental health prediction. Patient data, consisting of structured numerical assessments and unstructured therapist notes, is processed through a probabilistic routing mechanism. Depending on input characteristics, patients are assigned to specialized experts, single-task (STL) or multitask (MTL) models, optimized for numerical, textual, or combined modalities. This Mixture of Experts (MoE) approach dynamically adapts to patient heterogeneity, improving prediction accuracy and clinical relevance.}
    \label{fig:enter-label}
\end{figure}

\section{Experimental setup: synthetic multitask regression with adaptive routing}
\subsection{Synthetic data generation}
To evaluate our model's ability to learn input-dependent routing policies, we construct a synthetic multitask, multimodal regression benchmark. Each sample consists of two modalities: a numeric vector $\mathbf{x}^{(\mathrm{num})} \in \mathbb{R}^{d_{\mathrm{num}}}$ and a textual vector $\mathbf{x}^{(\mathrm{text})} \in \mathbb{R}^{d_{\mathrm{text}}}$. Both modalities contribute signal to both tasks, but their relevance varies across the input space.

To simulate this heterogeneity, we define input-dependent latent preferences over modality informativeness and task relevance. For each sample, we generate two outputs $y_1$ and $y_2$ corresponding to Task 1 and Task 2. These targets are nonlinear functions of both modalities, with task-specific coefficients and feature maps:
\begin{align*}
    y_1 &= \boldsymbol{\alpha}_1^\top \mathbf{x}^{(\mathrm{num})} + \boldsymbol{\beta}_1^\top \phi(\mathbf{x}^{(\mathrm{text})}) + \gamma_1 \cdot \sin(\boldsymbol{\omega}_1^\top \mathbf{x}^{(\mathrm{num})}) + \epsilon_1, \\
    y_2 &= \boldsymbol{\alpha}_2^\top \mathbf{x}^{(\mathrm{text})} + \boldsymbol{\beta}_2^\top \psi(\mathbf{x}^{(\mathrm{num})}) + \gamma_2 \cdot \cos(\boldsymbol{\omega}_2^\top \mathbf{x}^{(\mathrm{text})}) + \epsilon_2,
\end{align*}
where $\phi$ and $\psi$ are random Fourier feature maps and $\epsilon_k \sim \mathcal{N}(0, \sigma^2)$ is Gaussian noise. The parameters $\boldsymbol{\alpha}_k, \boldsymbol{\beta}_k, \gamma_k, \boldsymbol{\omega}_k$ are sampled independently for each trial. This setup introduces nonlinear, cross-modality interactions and supports fine-grained control over the input-task-modality relationships.

\begin{figure}[t]
    \centering
    \includegraphics[width=\textwidth]{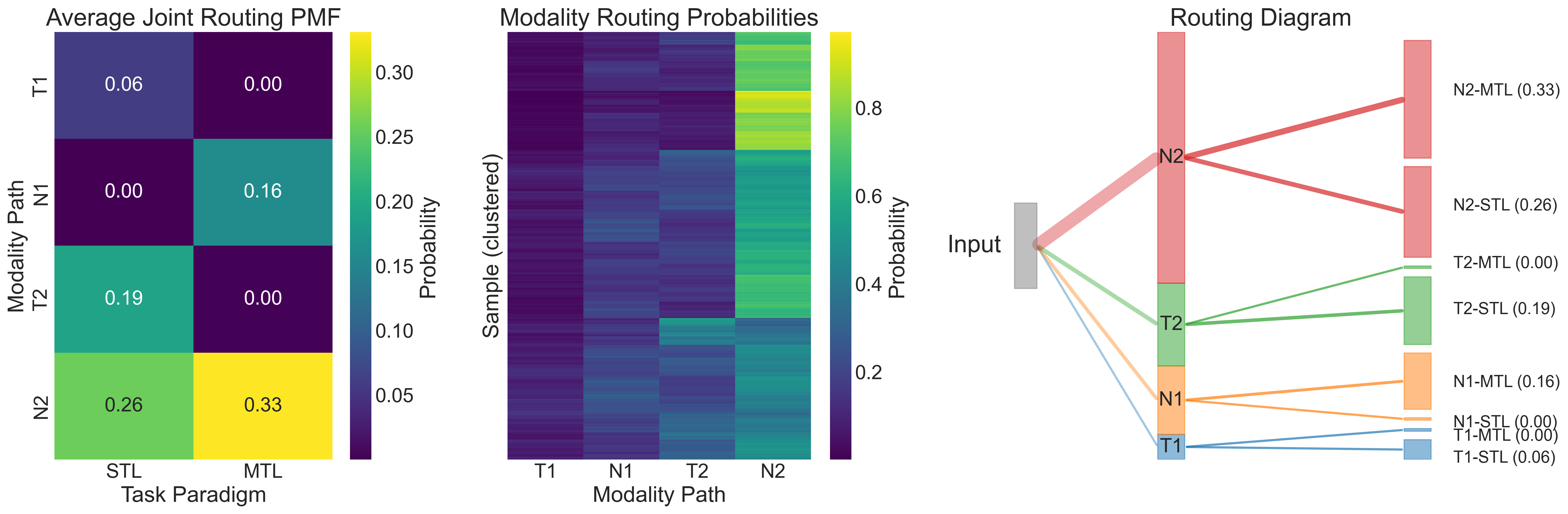}
    \caption{
\textbf{Routing behavior visualizations.}
(Left) \emph{Average Joint Routing PMF:} Heatmap showing average routing probabilities across all eight (modality, task paradigm) expert combinations. The model assigns most mass to N2-STL and N2-MTL, indicating a preference for fused numerical+text representations, especially under MTL.
(Middle) \emph{Modality Routing Probabilities:} Clustered heatmap of sample-specific routing distributions over four modality paths. The model segments the input space into distinct routing patterns, reflecting input-dependent computation.
(Right) \emph{Routing Diagram:} Sankey plot illustrating average routing policy. Edge widths are proportional to route probabilities. The architecture dynamically allocates across experts based on input characteristics.
}
    \label{fig:routing_summary}
\end{figure}

\begin{figure}[h]
    \centering
    \includegraphics[width=\linewidth]{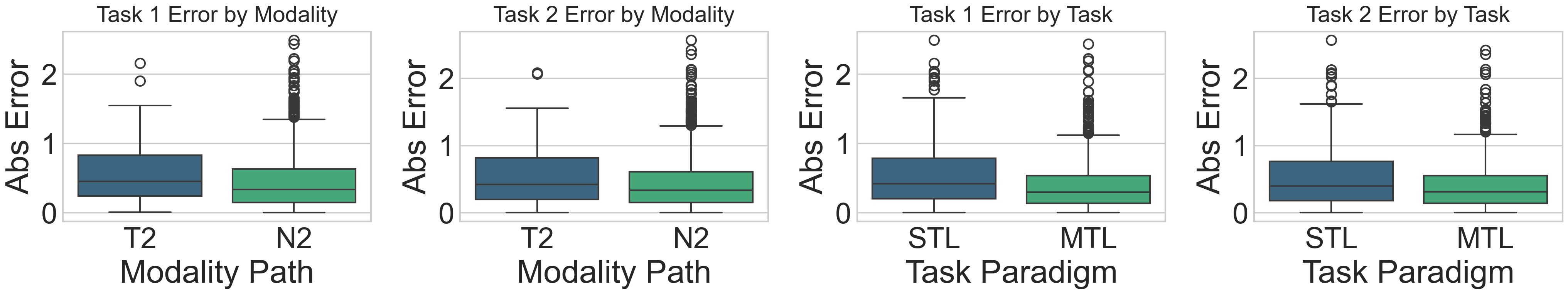}
    \caption{\textbf{Task-specific error distributions under hard routing.}
Boxplots show absolute prediction errors for each task, stratified by the selected modality path (T2, N2) and task paradigm (STL, MTL). The T2 and N2 configurations yield the lowest errors overall and dominate the hard routing assignments. Notably, T1 and N1 routes are effectively pruned by the router, receiving near-zero selection probability. MTL performs slightly better than STL, but not decisively enough to eliminate STL paths. These results demonstrate the benefit of adaptive routing across modalities and tasks.
}
\label{fig:task_error_hard}
\end{figure}

Taken together, Figures~\ref{fig:routing_summary} and~\ref{fig:task_error_hard} illustrate the model’s ability to align routing decisions with the latent structure of the task-modality landscape. Routes with higher selection probability correspond to lower predictive error, confirming that the learned policy not only adapts to input characteristics but also supports improved task performance. These results validate the effectiveness of probabilistic expert routing as a mechanism for uncovering and exploiting sample-specific patterns in multimodal multitask prediction.

\subsection{Real-world dataset: Mental Health in Healthcare Workers}
\begin{figure}[htb]
    \centering
\includegraphics[width=\textwidth]{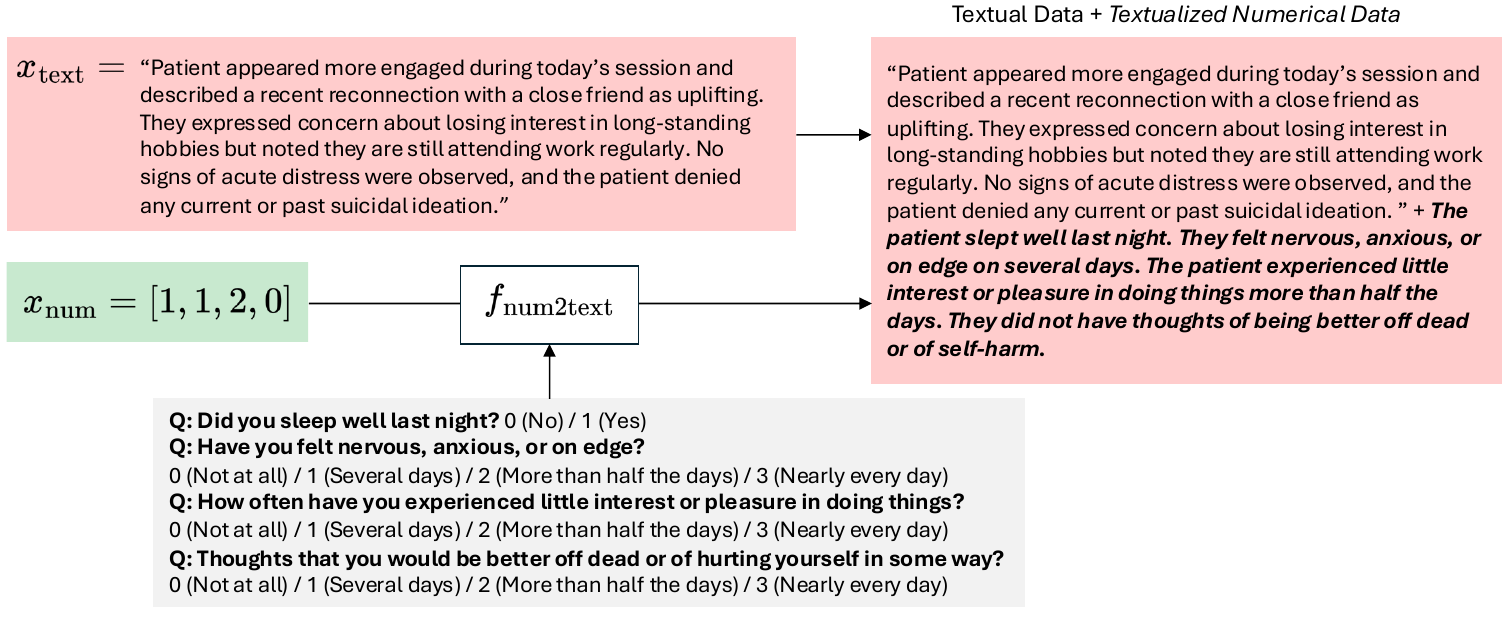}
    \caption{\textbf{Example of early fusion modality conversion.} (Left) Original data with numrical responses and therapist notes (not from a real patient to ensure privacy). (Right) Transformed representation where numerical responses are converted into text and concatenated with the original text, creating a unified textual modality.}
    \label{fig:translate}
\end{figure}

We applied our approach to an augmented real-world psychotherapy dataset involving healthcare workers (hereafter ``patients" for brevity) who presented with anxiety and/or depression symptoms during the course of the COVID-19 pandemic \citep{kanellopoulos2021copenyp, solomonov2022copenyp}. In this dataset, clinicians documented each session through unstructured notes (textual input, T1) and completed standardized clinical questionnaires (structured input, N1). Each patient was assessed using two outcome measures: depression severity (PHQ-9; Task 1) and anxiety severity (GAD-7; Task 2). To unify modalities, we transformed the structured numerical inputs into natural language using a fine-tuned text generation model, enabling early fusion with clinical notes to form a joint textual representation (T2). Figure~\ref{fig:translate} provides an example of this textualization process. Conversely, we encoded the unstructured clinical notes into numerical embeddings using a pretrained BERT model and concatenated these embeddings with the original structured features to construct a fused numerical representation (N2). This bidirectional transformation supports flexible modality routing and aligns heterogeneous inputs in a common representation. We evaluate using: (i) the predictive value of different modality paths, (ii) the benefit of multitask and heteroscedastic training objectives, and (iii) the effect of adaptive routing in selecting appropriate expert configurations.


Table \ref{tab:interpretation} summarizes the six expert configurations used in our model, derived from combinations of modality inputs (text, numerical, or both) and training objectives (STL or MTL). These correspond to the eight possible (modality, task) pathways in our mixture-of-experts routing scheme. Each configuration supports different patient profiles, for instance, patients with only clinical notes are routed to text-only experts, while patients with multiple structured assessments and correlated outcomes benefit more from MTL.

\begin{table}[htb!]
\caption{Overview of expert routes with their functions and ideal patient profiles.}
\label{tab:interpretation}
\centering
\footnotesize
\begin{tabular}{>{\centering\arraybackslash}p{0.2\textwidth} p{0.3\textwidth} p{0.5\textwidth}}
\toprule
\centering
\textbf{Route} & \textbf{Function} & \textbf{Interpretation} \\
\midrule

\textbf{Numerical (STL)} & 
Uses only structured scores (e.g., PHQ-9) to predict one outcome. & 
Patients with only standardized questionnaire scores (PHQ-9, GAD-7) and no, short, or noisy therapist notes. \\

\textbf{Numerical (MTL)} & 
Uses structured scores to predict multiple outcomes. & 
Patients with multiple structured scores available (PHQ-9, GAD-7) where outcomes are correlated. \\

\textbf{Text (STL)} & 
Uses only therapist notes to predict one outcome. & 
Patients without or noisy structured assessments but with detailed clinical notes with only one outcome available or two outcome with possible low correlation. \\

\textbf{Text (MTL)} & 
Uses only therapist notes to predict multiple outcomes. & 
Patients whose therapist notes describe multiple mental health problems (e.g., mentions of both depression and anxiety without or noisy structured scores). \\

\textbf{Numerical + Text (STL)} & 
Uses both structured and unstructured data but predicts only one outcome at a time. & 
Patients with both questionnaire scores and therapist notes with low outcome correlation. \\

\textbf{Numerical + Text (MTL)} & 
Uses both structured and unstructured data to predict multiple outcomes jointly. & 
Patients with both structured scores and clinical text, where multiple conditions (e.g., depression \& anxiety) need prediction. \\
\bottomrule

\end{tabular}
\end{table}

Table \ref{tab:compact_results} reports the predictive performance (RMSE) for each outcome. T1 (text-only) performs worst, particularly on PHQ-9, likely due to missing signal in free-text for structured outcomes. N1 (numerical-only) improves performance modestly. T2 (text + textualized numeric) achieves the best results overall, confirming that converting structured data into natural language and early-fusing it with notes enhances representation. N2 (numerical + embedded text) also performs well, but slightly under performs T2, suggesting limitations in treating text as static embeddings rather than contextual language input.

It also compares models trained with different objectives and routing strategies. STL under performs across both outcomes, especially PHQ-9, where inter-task signal is stronger. MTL improves over STL by learning shared representations. Introducing learned routing improves both tasks, confirming that adaptive pathway selection benefits generalization. Finally, combining routing and heteroscedastic loss yields the best performance overall, demonstrating that routing complements uncertainty-aware weighting. This model dynamically selects both modality and task structure on a per-sample basis and adjusts supervision strength based on estimated noise.

\begin{table}[htb]
\caption{RMSE on PHQ-9 / GAD-7 for all fixed-path baselines and model-level variants.}
\label{tab:compact_results}
\centering
\begin{tabular}{lcc}
\toprule
\textbf{Configuration} & \textbf{PHQ-9 RMSE} & \textbf{GAD-7 RMSE} \\
\midrule
\multicolumn{3}{l}{\emph{Modality-path baselines (STL)}} \\
\hspace{0.8em}T1  (Text-only)                    & 4.60 & 4.10 \\
\hspace{0.8em}N1  (Numerical-only)               & 4.08 & 3.85 \\
\hspace{0.8em}T2  (Textualized Num + Text)       & 3.66 & 3.42 \\
\hspace{0.8em}N2  (Num + Embedded Text)          & 3.80 & 3.58 \\[2pt]
\multicolumn{3}{l}{\emph{Training–objective / routing variants}} \\
\hspace{0.8em}STL (Independent)                  & 4.28 & 3.95 \\
\hspace{0.8em}MTL (Shared encoder)               & 4.12 & 3.78 \\
\hspace{0.8em}Heteroscedastic MTL               & 3.85 & 3.52 \\
\hspace{0.8em}Routing (Soft, learned paths)      & 3.62 & 3.34 \\
\bottomrule
\end{tabular}
\end{table}

\section{Discussion and Limitations}
\paragraph{Discussion.}
Our experiments show that an adaptive routing framework can improve multimodal multitask prediction under substantial input and task heterogeneity.
First, the synthetic benchmark confirms that the router discovers the latent mapping between modality relevance and task correlation. The model concentrates probability mass on paths that match the ground-truth data-generation rules and yields lower RMSE than static baselines.
Second, on the psychotherapy data, the framework selects the textualized-numeric path (T2) for samples with rich notes and numeric scores and prefers the numeric-only path (N1) when notes provide little extra signal. Routing probabilities align with clinical expectations: patients whose therapist notes give detailed symptom descriptions rely more on language models, while patients with minimal notes rely on structured scores.
Third, uncertainty-aware losses complement routing. The heteroscedastic objective discounts high-variance samples and reduces over-fitting, particularly for PHQ-9, which empirically exhibits greater label noise.
Further, the mixture-of-experts architecture yields interpretable sub-networks. Each expert specializes in a clearly defined modality–task configuration, which supports post-hoc inspection and potential deployment in clinical workflows where transparency is essential.  
 
\paragraph{Limitations.} The study relies on substantial synthetic data to stress-test routing behavior under controlled heterogeneity. Synthetic features follow idealized distributions, carry noise-free labels, and may embed patterns that rarely occur in real notes or questionnaire responses. Also, our “real-world” evaluation uses a hybrid corpus that augments genuine records with synthetically generated encounters, which expands data volume but creates distribution drift. Future work should validate on large untouched real-only test sets, apply importance weighting or domain-adversarial fine-tuning to reduce synthetic bias, and recalibrate class probabilities to reflect true clinical prevalence.

\begin{ack}
    This research is supported by funding from the National Institute of Aging (P30AG073105; a2Collective; Solomonov, Grosenick), the National Institute of Mental Health (R01MH131534; Grosenick)(K23 MH123864; Solomonov), the Cornell Center for Pandemic Prevention and Response (CCPPR) (Grosenick, Solomonov), and the Brain and Behavior Research Foundation (NARSAD Young Investigator Grant; Solomonov).
\end{ack}

\newpage
\bibliographystyle{apalike}
\bibliography{refs}

\newpage
\appendix

\section{Real-world data collection and usage}

\paragraph{Realistic Synthetic Data.}
The clinical outcomes described here were not used directly for model
training or evaluation in the main manuscript. Instead, they served as the basis of a highly realistic synthetic dataset that captures key distributional and structural characteristics of the real data. 

\paragraph{Ethical oversight.}
 Weill Cornell Medicine’s institutional review board approved the study; a waiver of informed consent was obtained for this retrospective study because it posed no more than minimal risk, did not affect care, rights or welfare and was deidentified for the purpose of analyses. 
 

\subsection*{Intervention and Data Collection}

\begin{itemize}
  \item \textbf{Context.} A four-session, remote, skills-based psychotherapy
        program was launched at a large academic medical center during the
        first peak of the COVID-19 pandemic (March 2020–April 2021).
  \item \textbf{Participants.} $N=534$ health care workers (HCWs)
        self-referred for emotional support. Role categories (percentages
        approximated): 35.2\% nursing, 24.3\% patient support, 22.8\%
        administrative, 13.8\% medical trainees/faculty, 2.4\% facilities,
        1.3\% family members. 70\% were on-site (frontline); 19\% worked
        remotely; 11\% unspecified.
  \item \textbf{Clinicians.} Sixty-seven trained providers (licensed
        psychologists, psychiatrists, social workers, and supervised trainees)
        delivered a total of 1,423 telehealth sessions.
  \item \textbf{Measures.}
        \begin{itemize}
          \item Patient Health Questionnaire–9 (PHQ-9) and Generalized Anxiety
                Disorder–7 (GAD-7) at sessions 1 and 4.
          \item PHQ-4 at sessions 2 and 3 for interim symptom tracking.
          \item Columbia Suicide Severity Rating Scale (C-SSRS) at intake and
                as needed for suicide risk.
        \end{itemize}
  \item \textbf{Safety.} Participants with severe symptoms or safety concerns
        were referred to appropriate emergency or long-term psychiatric care.
\end{itemize}
A linear mixed-effects model (random intercept and slope per subject; fixed
effect of time) showed significant reduction in overall symptom burden:
\begin{align*}
\text{PHQ-4}_{\text{baseline}} &= 5.65 \pm 2.95 
\;\longrightarrow\; 
\text{PHQ-4}_{\text{final}} = 3.32 \pm 2.46, \\
F_{3,823} &= 109.23,\; p < .001,\; 
\eta^2_{\text{partial}} = 0.27.
\end{align*}

For participants with clinically elevated symptoms at baseline
(PHQ-4 $\ge 6$), the effect size was stronger
($\eta^2_{\text{partial}} = 0.46$). Response rates were 42\% on GAD-7 and 43\%
on PHQ-9 ($ \geq50\%$ symptom reduction).

\section{Data-Driven (Empirical) Synthetic Data}

To support controlled experimentation and model probing, we generated synthetic samples that reflect “clean” examples from the original dataset. These samples were designed to preserve strong predictive signal across both GAD and PHQ outcomes. The process is structured to maintain cross-modal coherence between structured numerical features and free-text clinical notes. The generation process proceeds in three stages: (1)	Sampling structured numerical features from a smoothed approximation of high-confidence empirical distributions. (2)	Generating text conditionally based on these numerical features using a large language model (LLM) with custom prompting. (3) Filtering generated samples using our trained multimodal multitask model to retain only those with low predictive uncertainty and high task-specific confidence.

\subsection{Synthetic Numerical Data Generation}

We generated synthetic numerical data using three methods, each designed to preserve the statistical structure of the original dataset.

\textbf{Gaussian Synthesis.}
We estimated the mean vector \( \boldsymbol{\mu} \in \mathbb{R}^d \) and covariance matrix \( \boldsymbol{\Sigma} \in \mathbb{R}^{d \times d} \) from the original dataset. To ensure numerical stability, a small constant \( 10^{-6} \) was added to the diagonal of \( \boldsymbol{\Sigma} \). Synthetic samples were then drawn from a multivariate normal distribution:
\[
\mathbf{X}_{\text{synthetic}} \sim \mathcal{N}(\boldsymbol{\mu}, \boldsymbol{\Sigma}),
\quad \mathbf{X}_{\text{synthetic}} \in \mathbb{R}^{n_{\text{samples}} \times d}.
\]

\textbf{Copula-Based Synthesis.}
We applied a rank-based transformation to map each feature to a standard normal distribution. The empirical correlation matrix was computed on the transformed data. Synthetic samples were drawn from a multivariate normal distribution with this correlation structure and subsequently mapped back to the original marginal distributions using a quantile transform. This method preserved nonlinear dependencies among features.

\textbf{Kernel Density Estimation (KDE) Synthesis.}
Continuous features were standardized using \texttt{StandardScaler}. A Gaussian kernel density estimator was fitted with bandwidth \( h = n^{-1/(d+4)} \), where \( n \) is the number of samples and \( d \) is the feature dimensionality. New samples were drawn using KDE-based resampling. Binary features (e.g., \textit{dissociate}, \textit{anger}, \textit{fear\_contam}) were thresholded post-generation:
\[
\mathbf{X}_{\text{synthetic}}[\text{binary}] = (\mathbf{X}_{\text{synthetic}}[\text{binary}] > 0.5)\texttt{.astype(int)}.
\]

For each method, we generated 200 synthetic samples while maintaining the original class distribution of PHQ-9 and GAD-7 binary outcomes. The synthetic datasets closely matched the original statistical characteristics, with a mean absolute difference in feature correlation of less than 0.1 and a KL divergence in class distribution of less than 0.05.

\subsection*{Synthetic Text Data Generation}

To generate synthetic text data that reflect realistic psychotherapy session notes, we employed a prompt-based generation pipeline using small, open-source large language models (LLMs). The goal was to create natural language samples that are consistent with the patterns observed in the original dataset, while maintaining data privacy and avoiding memorization of sensitive content.

\textbf{Numerical-to-Text Conversion.}
We first transformed structured numerical features into short natural language descriptions. For each synthetic numerical instance \( x \in \mathbb{R}^d \), we constructed a template-based summary containing key symptom indicators and severity scores (e.g., PHQ-9, GAD-7). An example of this intermediate representation is:
\begin{quote}
\textit{The patient reported a PHQ-9 score of 15 and a GAD-7 score of 13. They endorsed symptoms such as dissociation and irritability, with no signs of fear of contamination.}
\end{quote}

\textbf{LLM-Based Natural Language Expansion.}
We used an instruction-tuned language model (Flan-T5 or Phi-2) to expand the structured summaries into fluent and contextually appropriate psychotherapy notes. Each prompt followed the format:
\begin{quote}
\textit{Patient data: PHQ-9 = 15, GAD-7 = 13, symptoms = [dissociation, irritability]. Write a brief therapist note summarizing the patient’s emotional state and challenges.}
\end{quote}
The model generated coherent text such as:
\begin{quote}
\textit{The patient presented with moderate symptoms of depression and anxiety, including dissociative experiences and heightened irritability. They expressed difficulty managing emotional stressors and reported low energy and trouble sleeping.}
\end{quote}

\textbf{Post-Processing and Filtering.}
We generated one therapist-style note per synthetic numerical input, resulting in 200 synthetic text samples. To ensure linguistic diversity and clinical plausibility, we applied basic heuristics to filter out degenerate outputs (e.g., overly repetitive or off-topic content). The vocabulary and sentence structure were qualitatively consistent with those in the original data, and generated texts maintained semantic alignment with the associated synthetic numerical features.

\textbf{Privacy Considerations.}
All models were run locally without external API calls to ensure HIPAA compliance. We used only open-access, instruction-tuned models with small memory footprints (Phi-2), which allowed controlled offline generation and ensured that no real patient data were exposed or used during synthesis.

\begin{figure}[h]
    \centering
    \includegraphics[width=\linewidth]{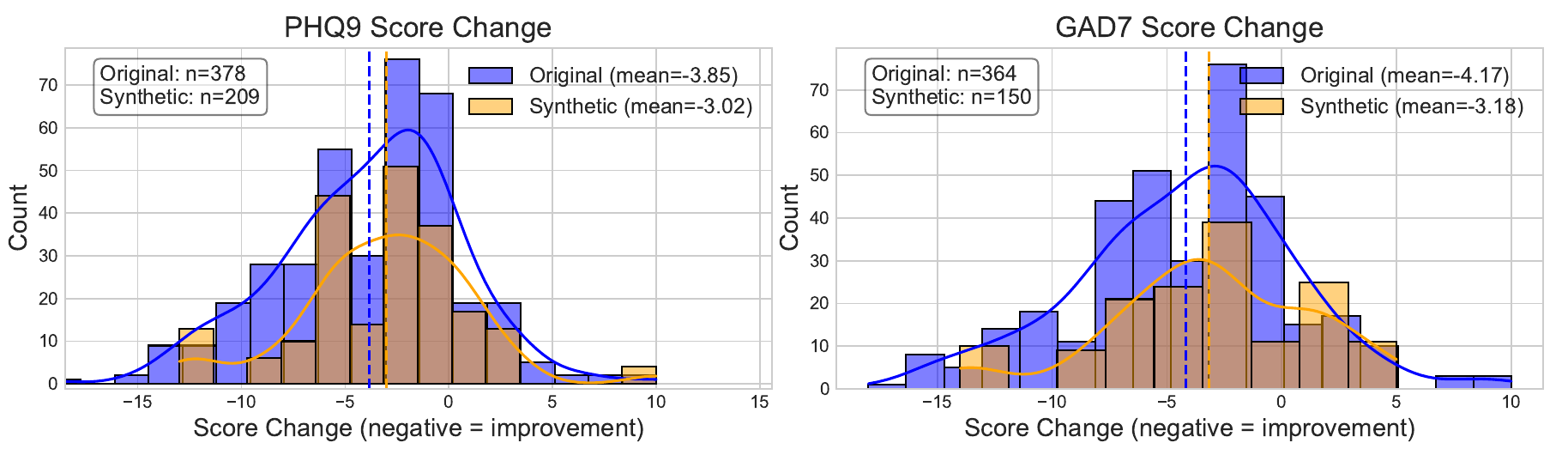}
    \caption{Distribution of synthetic and original data}
    \label{fig:syn-dis}
\end{figure}

\section{Equation-Driven (Analytical) Synthetic Data}
Our synthetic benchmark is designed to rigorously test input-dependent routing in multitask, multimodal regression. Each sample consists of two modalities, a numeric vector and a text vector, and two regression targets. Both modalities contribute to both tasks, but the degree and nature of their informativeness is heterogeneous and input-dependent, simulating real-world complexity.

\paragraph{Input Features.}
For each sample, we generate:
\begin{itemize}
    \item Numeric features $\mathbf{x}^{(\mathrm{num})} \in \mathbb{R}^{d_{\mathrm{num}}}$, sampled as $\mathcal{N}(0, I)$.
    \item Text features $\mathbf{x}^{(\mathrm{text})} \in \mathbb{R}^{d_{\mathrm{text}}}$, also sampled as $\mathcal{N}(0, I)$.
\end{itemize}
We use $d_{\mathrm{num}} = d_{\mathrm{text}} = 16$ unless otherwise specified.

\paragraph{Random Fourier Feature Maps.}
To introduce nonlinear, cross-modal dependencies, we use random Fourier feature (RFF) maps:
\begin{align*}
    \phi(\mathbf{x}^{(\mathrm{text})}) &= \sqrt{\frac{2}{D}} \cos(W_\phi \mathbf{x}^{(\mathrm{text})} + b_\phi), \\
    \psi(\mathbf{x}^{(\mathrm{num})}) &= \sqrt{\frac{2}{D}} \cos(W_\psi \mathbf{x}^{(\mathrm{num})} + b_\psi),
\end{align*}
where $W_\phi, W_\psi \in \mathbb{R}^{D \times 16}$ have entries drawn from $\mathcal{N}(0, 1)$, $b_\phi, b_\psi \in \mathbb{R}^D$ are drawn uniformly from $[0, 2\pi]$, and $D=32$.

\paragraph{Target Construction.}
For each sample, we generate two targets:
\begin{align*}
    y_1 &= \boldsymbol{\alpha}_1^\top \mathbf{x}^{(\mathrm{num})} + \boldsymbol{\beta}_1^\top \phi(\mathbf{x}^{(\mathrm{text})}) + \gamma_1 \cdot \sin(\boldsymbol{\omega}_1^\top \mathbf{x}^{(\mathrm{num})}) + \epsilon_1, \\
    y_2 &= \boldsymbol{\alpha}_2^\top \mathbf{x}^{(\mathrm{text})} + \boldsymbol{\beta}_2^\top \psi(\mathbf{x}^{(\mathrm{num})}) + \gamma_2 \cdot \cos(\boldsymbol{\omega}_2^\top \mathbf{x}^{(\mathrm{text})}) + \epsilon_2,
\end{align*}
where:
\begin{itemize}
    \item $\boldsymbol{\alpha}_k, \boldsymbol{\beta}_k \sim \mathcal{N}(0, I)$ (dimensions match their arguments)
    \item $\boldsymbol{\omega}_k \sim \mathcal{N}(0, I)$ (dimension 16)
    \item $\gamma_k \sim \mathrm{Uniform}[0.5, 1.5]$
    \item $\epsilon_k \sim \mathcal{N}(0, 0.1^2).$
\end{itemize}
All parameters are independently sampled for each trial, and fixed for all samples within a trial.

\paragraph{Design Rationale.}
This construction ensures:
\begin{itemize}
    \item Both modalities are relevant to both tasks, but in different, nonlinear, and cross-modal ways.
    \item The use of RFFs simulates learned embeddings and increases the complexity of the mapping.
    \item Sinusoidal nonlinearities further challenge the model, requiring it to capture nontrivial dependencies.
    \item The setup allows for controlled ablations (e.g., by zeroing coefficients or removing nonlinear terms).
\end{itemize}

\paragraph{Implementation Notes.}
\begin{itemize}
    \item All random seeds are fixed for reproducibility.
    \item The code for data generation is provided in the supplementary repository.
    \item The synthetic dataset can be easily extended to more modalities or tasks by following the same recipe.
\end{itemize}

\begin{algorithm}[H]
\caption{Synthetic Data Generation}
\begin{algorithmic}[1]
\STATE Sample $\mathbf{x}^{(\mathrm{num})}, \mathbf{x}^{(\mathrm{text})} \sim \mathcal{N}(0, I_{16})$
\STATE Compute $\phi(\mathbf{x}^{(\mathrm{text})})$, $\psi(\mathbf{x}^{(\mathrm{num})})$ via RFFs
\STATE Sample $\boldsymbol{\alpha}_k, \boldsymbol{\beta}_k, \boldsymbol{\omega}_k, \gamma_k$ as above
\STATE Compute $y_1, y_2$ as above, add noise $\epsilon_k$
\end{algorithmic}
\end{algorithm}

\begin{verbatim}
def rff(x, W, b):
    return np.sqrt(2 / W.shape[0]) * np.cos(W @ x + b)

x_num = np.random.randn(16)
x_text = np.random.randn(16)
W_phi, b_phi = np.random.randn(32, 16), np.random.uniform(0, 2*np.pi, 32)
W_psi, b_psi = np.random.randn(32, 16), np.random.uniform(0, 2*np.pi, 32)
phi_x_text = rff(x_text, W_phi, b_phi)
psi_x_num = rff(x_num, W_psi, b_psi)
# ... sample coefficients and compute y1, y2 as above
\end{verbatim}

\subsection{Scenario 1: Sinusoidal/Cosine, Both Modalities}
\begin{verbatim}
Number of samples: 1000 (train), 1000 (test)
Feature dimensions: d_num = 16, d_text = 16, D (RFF output) = 32

Feature maps:
phi(x_text) = sqrt(2/32) * cos(W_phi x_text + b_phi)
psi(x_num) = sqrt(2/32) * cos(W_psi x_num + b_psi)
W_phi, W_psi ~ N(0, 1), b_phi, b_psi ~ Uniform[0, 2pi]

Target equations:
y1 = alpha1^T x_num + beta1^T phi(x_text) + gamma1 * sin(omega1^T x_num) + epsilon1
y2 = alpha2^T x_text + beta2^T psi(x_num) + gamma2 * cos(omega2^T x_text) + epsilon2

Parameter distributions:
- alpha_k, beta_k ~ N(0, I)
- omega_k ~ N(0, I)
- gamma_k ~ Uniform[0.5, 1.5]
- epsilon_k ~ N(0, 0.1^2)
\end{verbatim}

\begin{figure}
    \subfloat[]{\includegraphics[width=\linewidth]{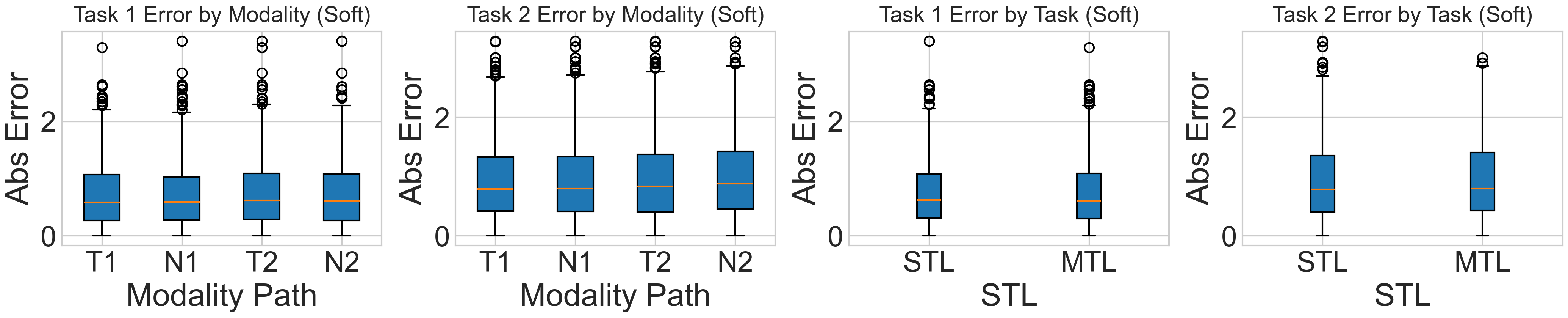}}\\
    \subfloat[]{\includegraphics[width=\linewidth]{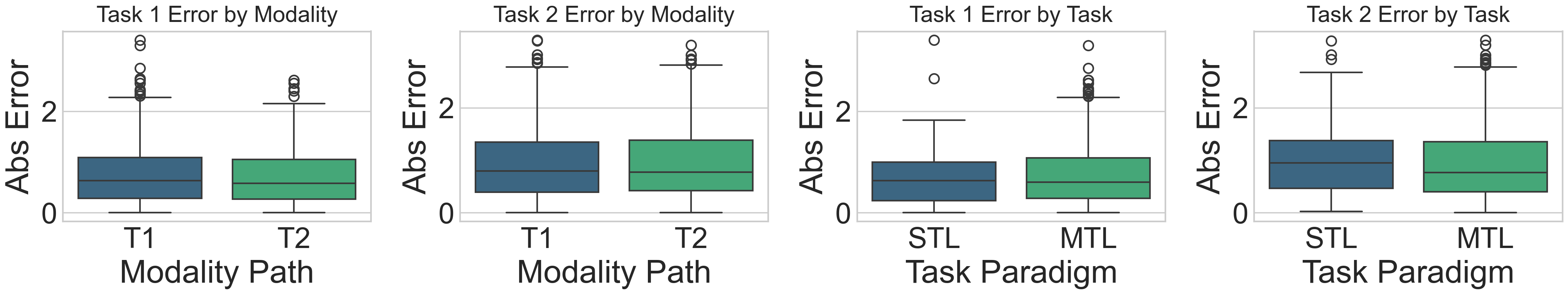}}\\
    \subfloat[]{    \includegraphics[width=\linewidth]{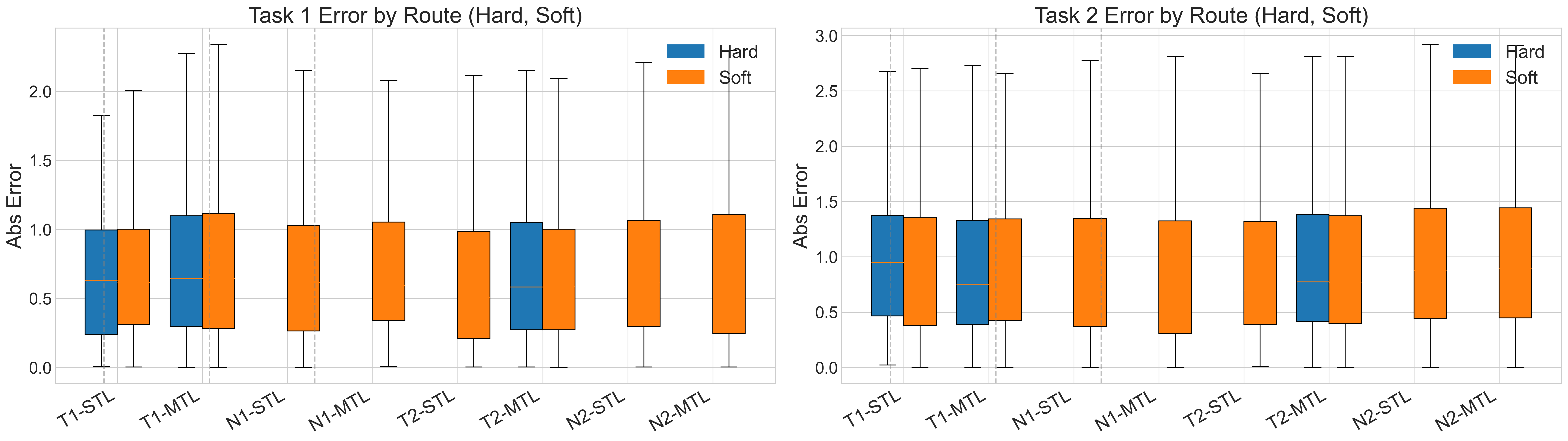}}\\
    \subfloat[]{    \includegraphics[width=\linewidth]{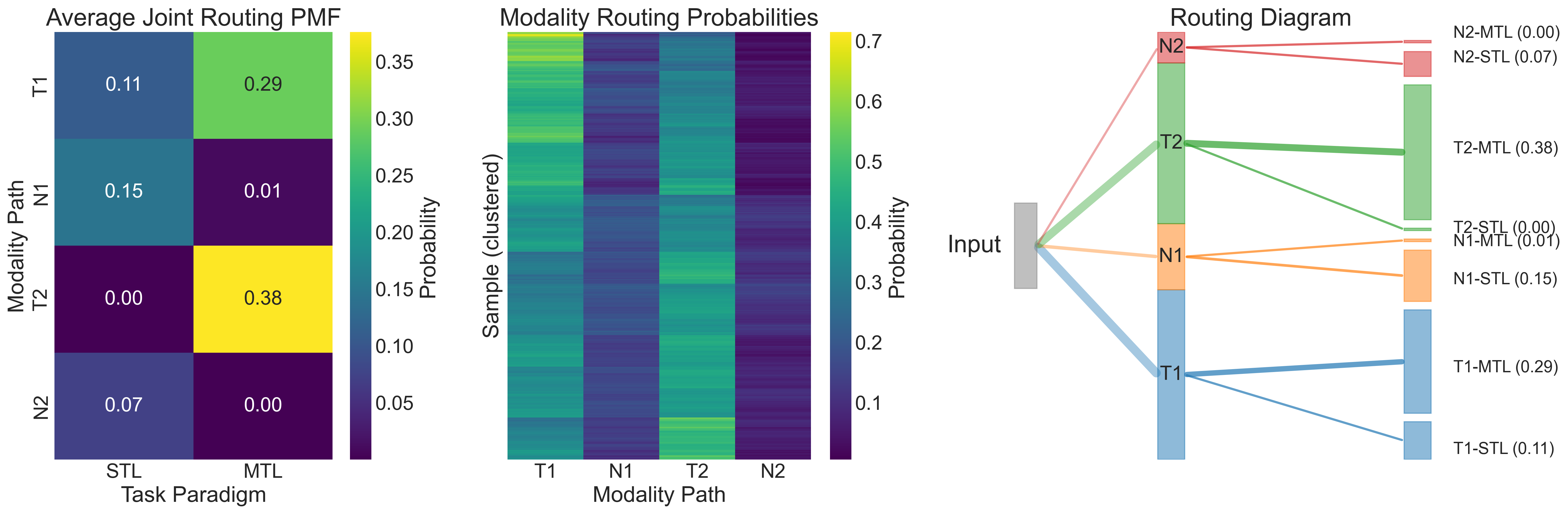}}\\
    \subfloat[]{    \includegraphics[width=0.8\linewidth]{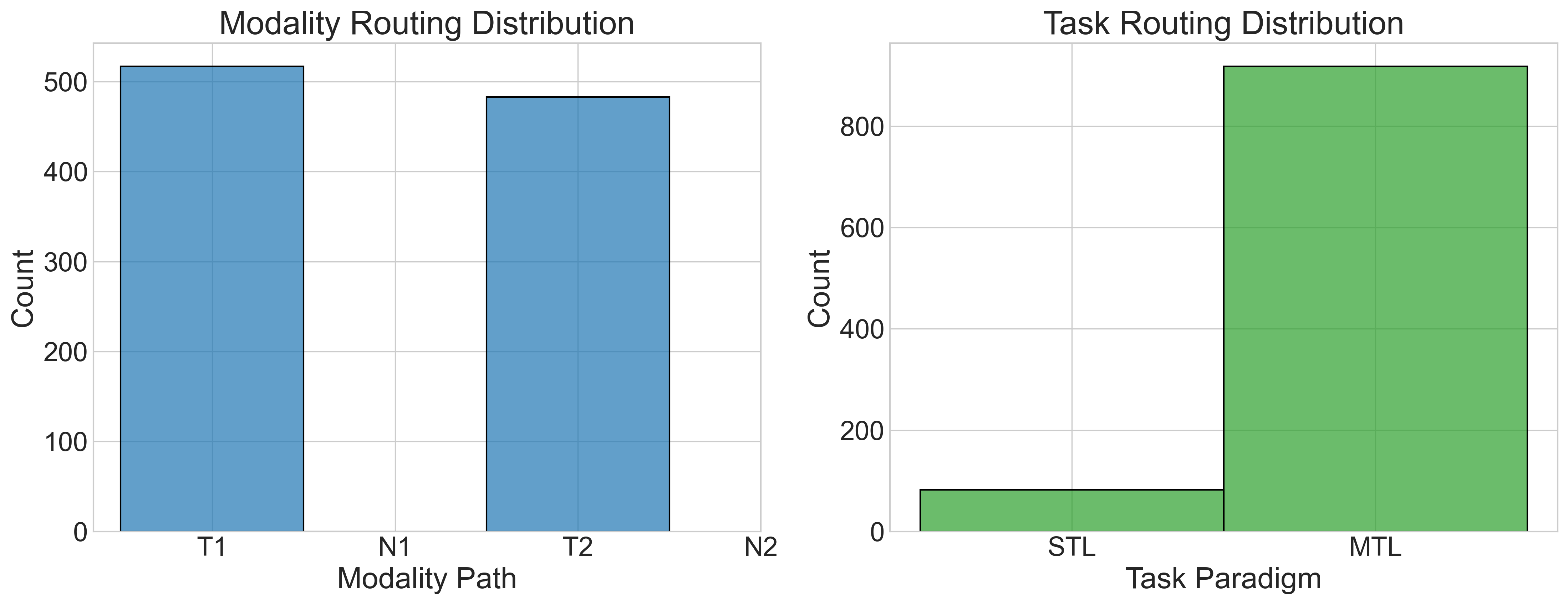}}
    \caption{ \textbf{Scenario 1: Sinusoidal/Cosine, Both Modalities}
(a) Absolute error by route (soft routing, weighted by probabilities).
(b) Absolute error by route (hard routing, based on most probable path).
(c) Comparison of absolute errors for each route: hard vs. soft routing (note that in hard routing, boxes do not appear in every case as the router can learn to bypass N1 as a less probable path).
(d) Summary: joint routing PMF, clustered heatmap, and routing Sankey diagram.
(e) Distribution of selected modality and task paradigm by the router.
}
\label{fig:sc1}
\end{figure}

In this scenario, as expected, the model learns that using MTL yields the best performance (see Figure \ref{fig:sc1}).

\subsection{Scenario 2: STL Preferred}
\begin{verbatim}
Number of samples: 1000 (train), 1000 (test)
Feature dimensions: d_num = 16, d_text = 16, D (RFF output) = 32

Feature maps:
phi(x_text) = sqrt(2/32) * cos(W_phi x_text + b_phi)
psi(x_num) = sqrt(2/32) * cos(W_psi x_num + b_psi)
W_phi, W_psi ~ N(0, 1), b_phi, b_psi ~ Uniform[0, 2pi]

Target equations:
y1 = alpha1^T x_num + gamma1 * sin(omega1^T x_num) + epsilon1
y2 = alpha2^T x_text + gamma2 * cos(omega2^T x_text) + epsilon2

Parameter distributions:
- alpha_k ~ N(0, I)
- omega_k ~ N(0, I)
- gamma_k ~ Uniform[0.5, 1.5]
- epsilon_k ~ N(0, 0.1^2)
\end{verbatim}

\begin{figure}[h]
    \subfloat[]{\includegraphics[width=\linewidth]{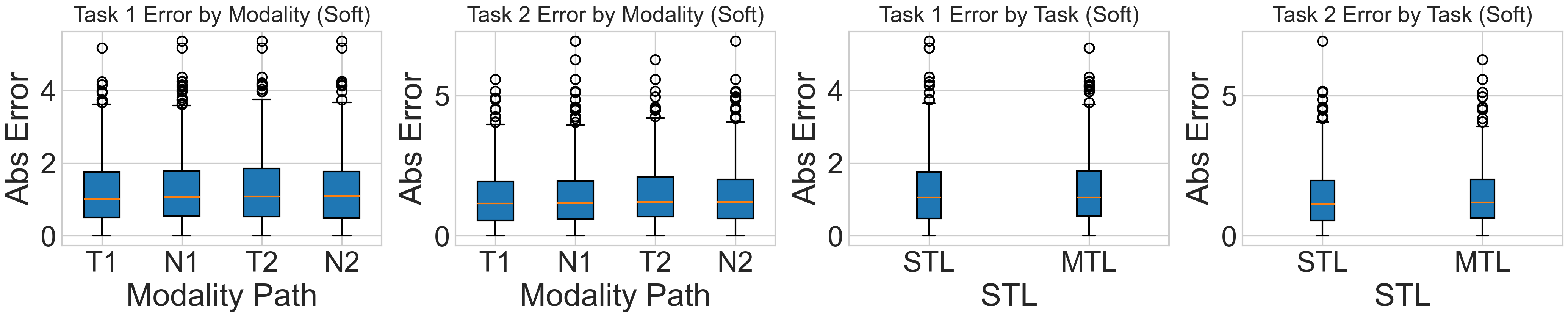}}\\
    \subfloat[]{\includegraphics[width=\linewidth]{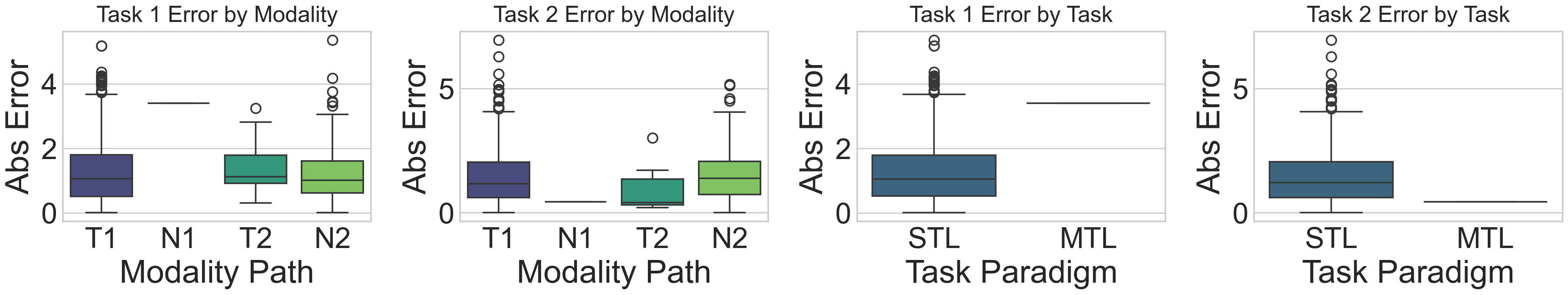}}\\
    \subfloat[]{    \includegraphics[width=\linewidth]{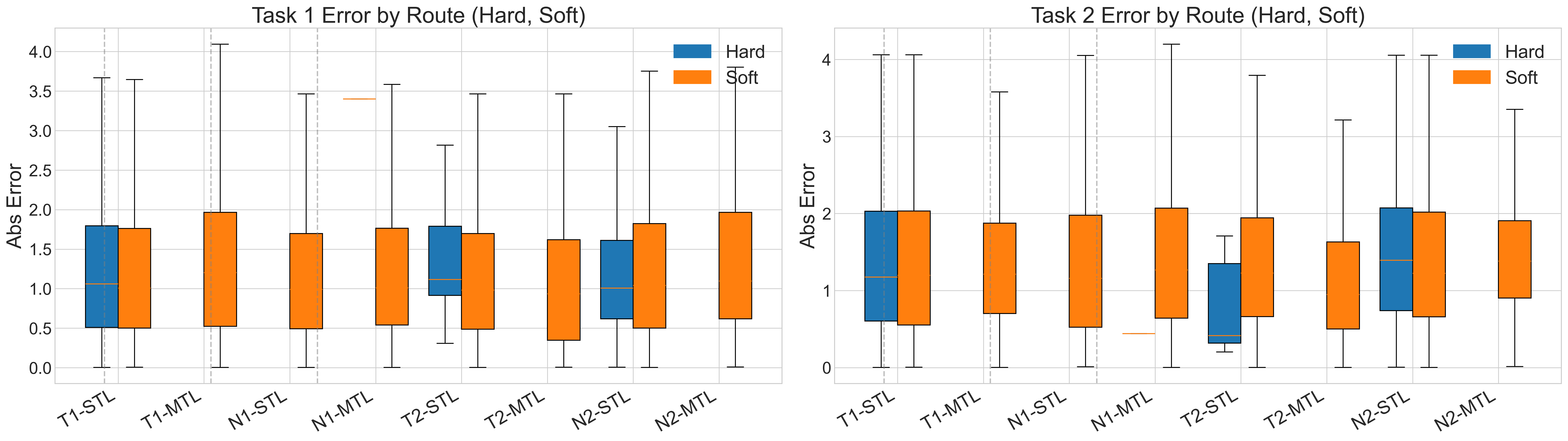}}\\
    \subfloat[]{    \includegraphics[width=\linewidth]{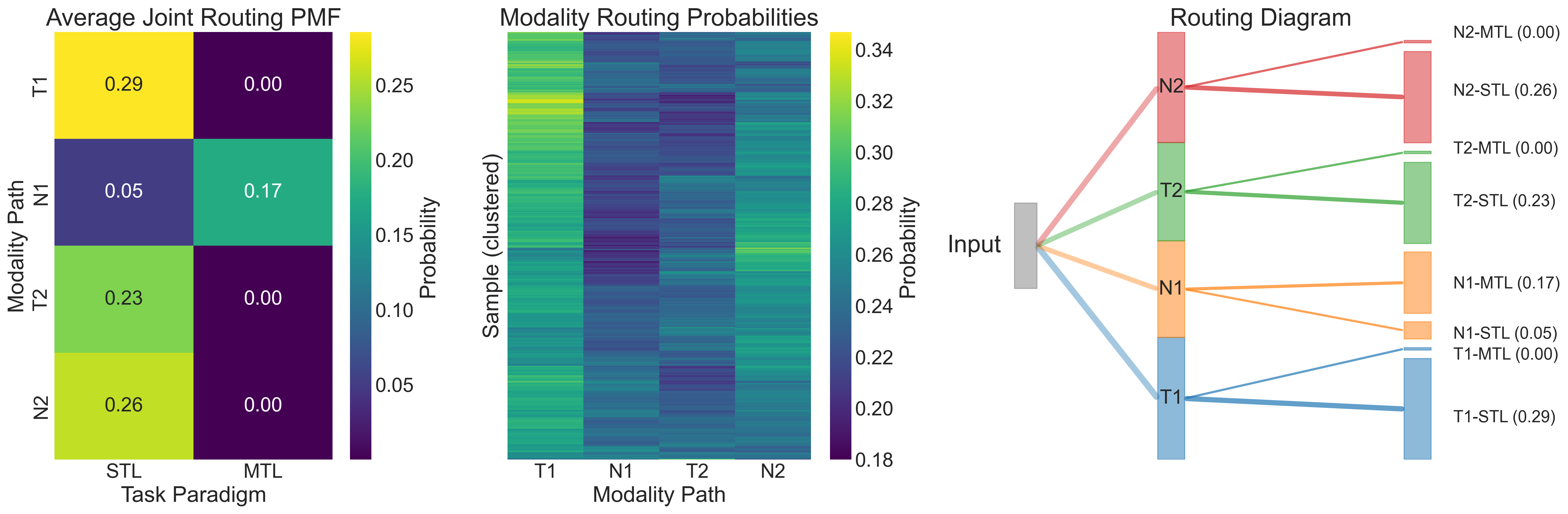}}\\
    \subfloat[]{    \includegraphics[width=0.8\linewidth]{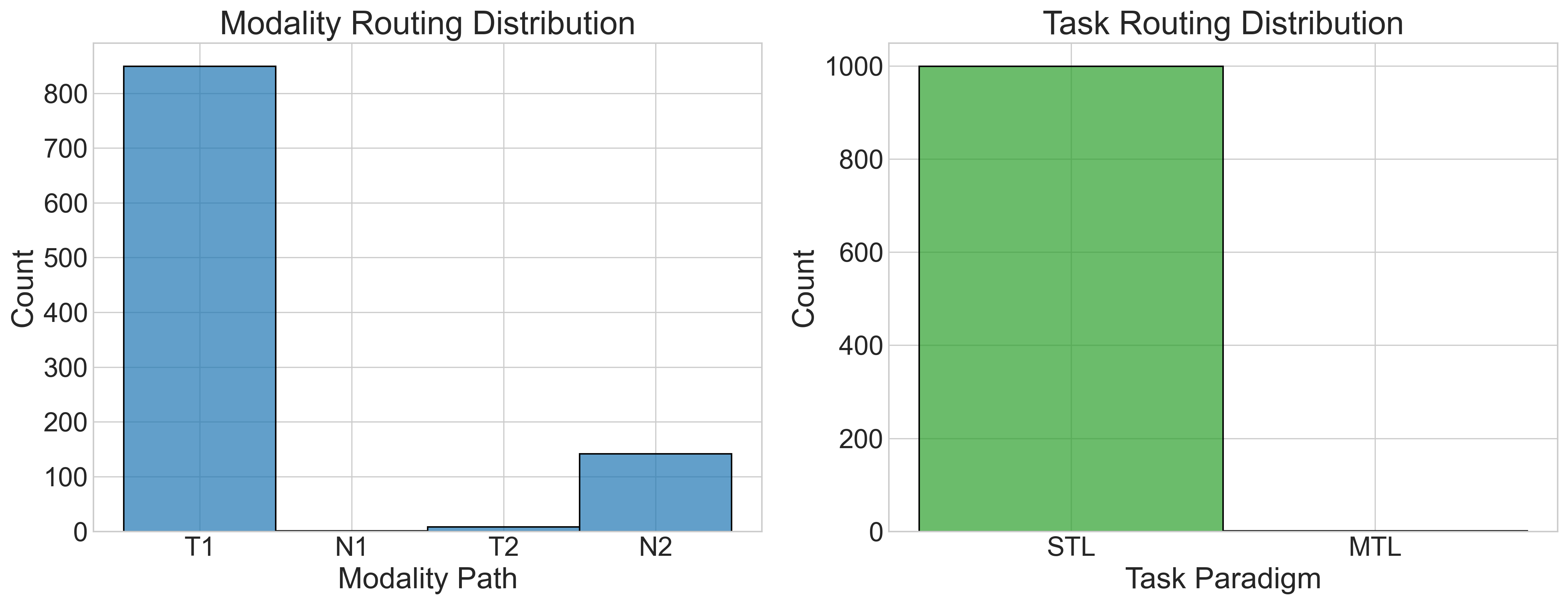}
    }
    \caption{ \textbf{Scenario 2: STL Prefered}
(a) Absolute error by route (soft routing, weighted by probabilities).
(b) Absolute error by route (hard routing, based on most probable path).
(c) Comparison of absolute errors for each route: hard vs. soft routing (note that in hard routing, boxes do not appear in every case).
(d) Summary: joint routing PMF, clustered heatmap, and routing Sankey diagram.
(e) Distribution of selected modality and task paradigm by the router.
}
\label{fig:sc2}
\end{figure}

In this scenario, each task is generated from a single modality: $y_1$ depends only on the numeric features and their nonlinear transformation, while  $y_2$ depends only on the textual features and their nonlinear transformation. Since each task is generated independently from its own modality, there is no shared information or benefit to learning the tasks jointly. As a result, the model learns that treating each task separately by using STL yields the best performance (see Figure \ref{fig:sc2}).

\subsection{Scenario 3: Fusion-Dominant Routing}
\begin{verbatim}
Number of samples: 1000 (train), 1000 (test)
Feature dimensions: d_num = 16, d_text = 16, D (RFF output) = 32

Feature maps:
phi(x_text) = sqrt(2/32) * cos(W_phi x_text + b_phi)
psi(x_num) = sqrt(2/32) * cos(W_psi x_num + b_psi)
W_phi, W_psi ~ N(0, 1), b_phi, b_psi ~ Uniform[0, 2pi]

Target equations:
y1 = alpha1^T x_num + beta1^T phi(x_text)  + epsilon1
y2 = alpha2^T x_text + beta2^T psi(x_num)  + epsilon2

Parameter distributions:
- alpha_k, beta_k ~ N(0, I)
- omega_k ~ N(0, I)
- epsilon_k ~ N(0, 0.1^2)
\end{verbatim}

\begin{figure}[h]
    \subfloat[]{\includegraphics[width=\linewidth]{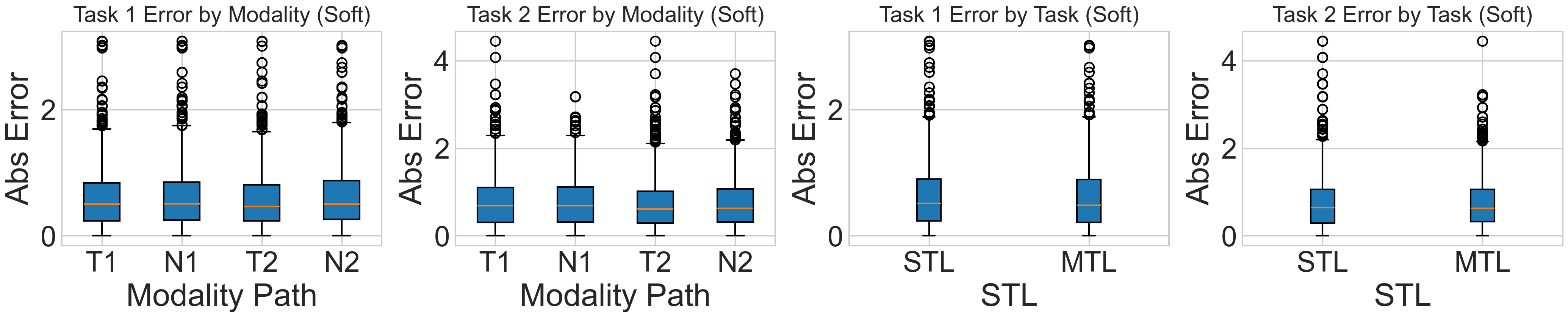}}\\
    \subfloat[]{\includegraphics[width=\linewidth]{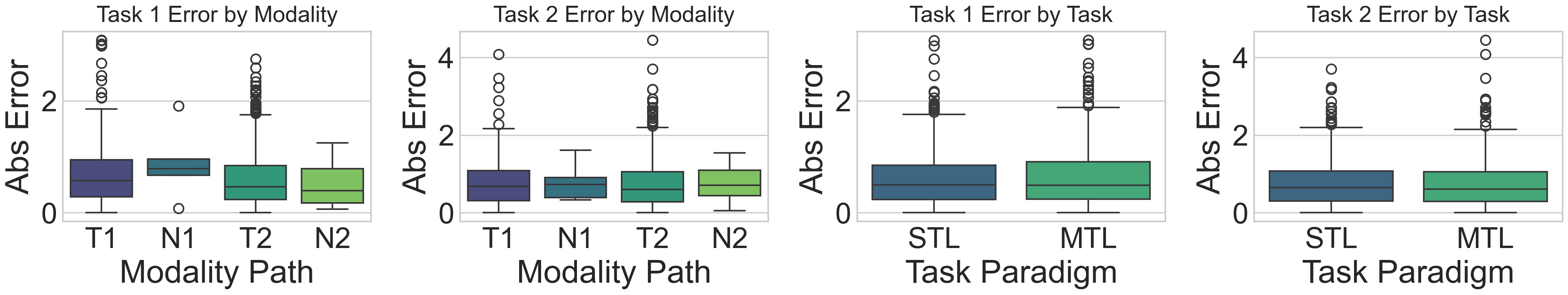}}\\
    \subfloat[]{    \includegraphics[width=\linewidth]{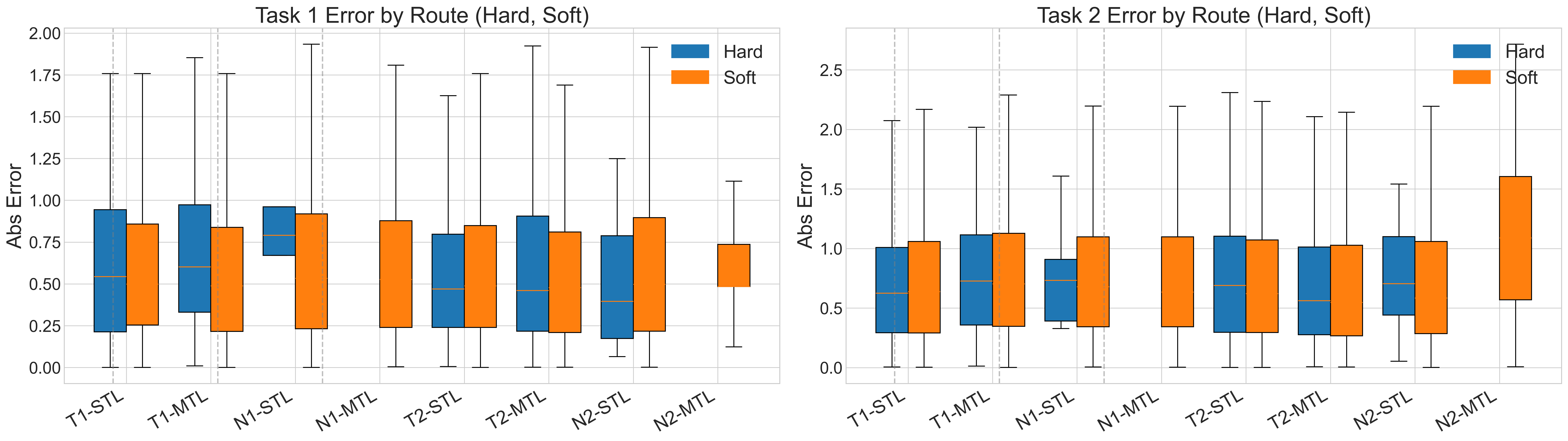}}\\
    \subfloat[]{    \includegraphics[width=\linewidth]{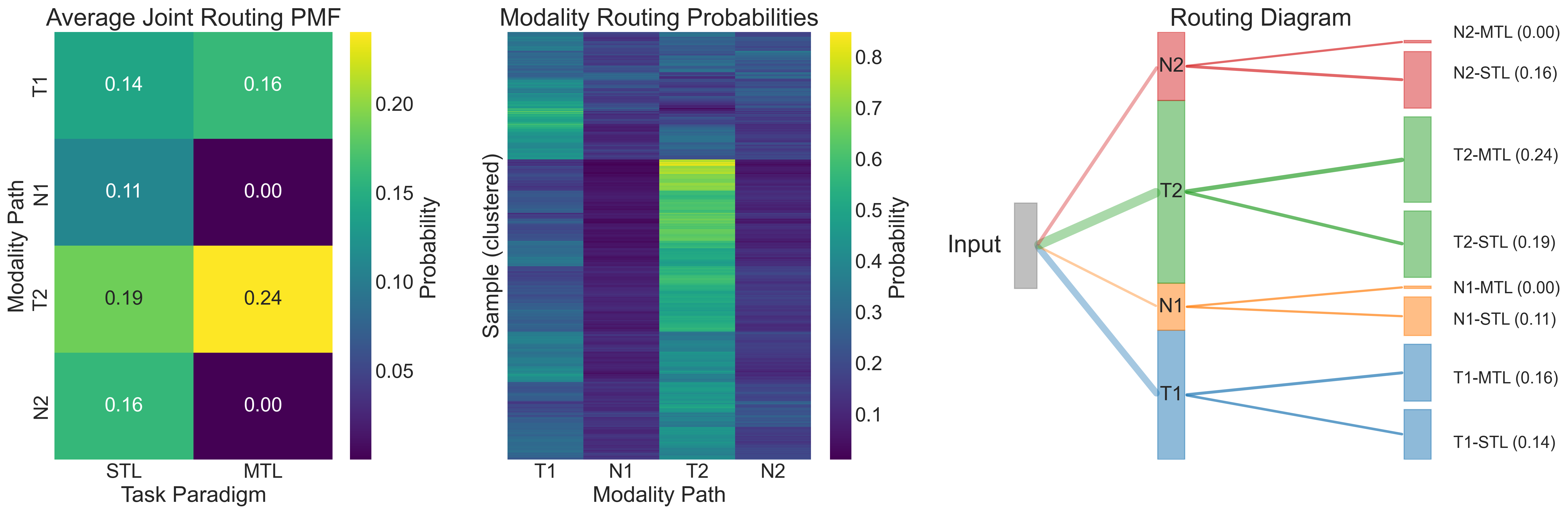}}\\
    \subfloat[]{    \includegraphics[width=0.8\linewidth]{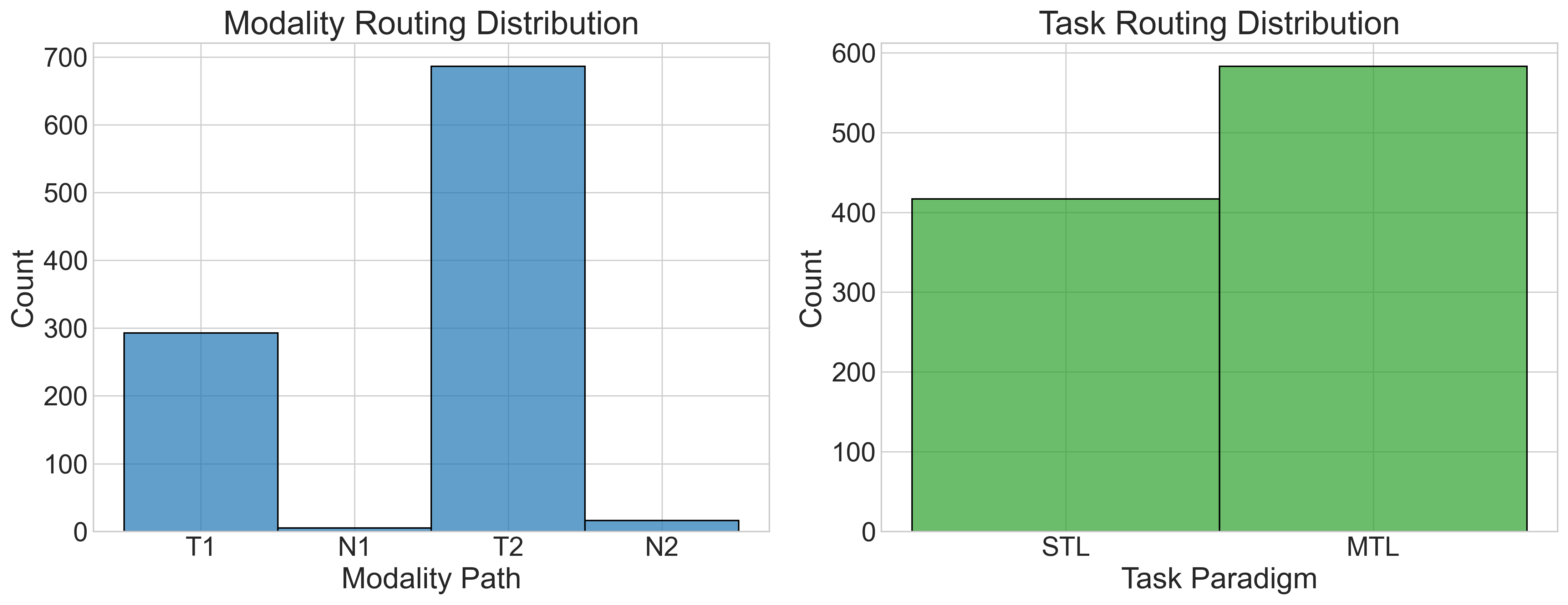}
    }
    \caption{ \textbf{Scenario 3: Fusion-Dominant Routing}
(a) Absolute error by route (soft routing, weighted by probabilities).
(b) Absolute error by route (hard routing, based on most probable path).
(c) Comparison of absolute errors for each route: hard vs. soft routing (note that in hard routing, boxes do not appear in every case).
(d) Summary: joint routing PMF, clustered heatmap, and routing Sankey diagram.
(e) Distribution of selected modality and task paradigm by the router.
}
\label{fig:sc3}
\end{figure}

The results in Figure \ref{fig:sc3}show that the model prefers the T2 (fusion) modality path, especially in combination with the MTL (multi-task learning) paradigm. The N1 and N2 paths are rarely or never used for MTL, indicating that the model has learned to avoid these routes in favor of more effective ones.

\section{Broader Impact}

This work introduces a flexible machine learning framework for adaptively routing data through multimodal and multitask pathways, with primary application to psychological outcome prediction. By enabling models to select personalized computation paths based on both input availability and task structure, this approach has the potential to improve prediction accuracy and robustness in real-world settings where data is heterogeneous and incomplete.

While our evaluation is framed in the context of mental health prediction, the methodology is broadly applicable to domains such as clinical decision support, education, and human-centered AI systems where structured and unstructured inputs coexist and multiple outcomes must be considered jointly.

At the same time, predictive models in healthcare and mental health raise significant ethical concerns. These include the risk of reinforcing biases present in clinical documentation, the opacity of model decisions, and the potential for overreliance on algorithmic outputs in high-stakes scenarios. Our model attempts to mitigate some of these risks by producing interpretable routing decisions, which may offer insight into modality usefulness and task interactions. Nonetheless, interpretability and fairness should be further studied before deployment.

Our work uses synthetic data and carefully preprocessed clinical data to demonstrate technical contributions, and does not aim to inform clinical decisions directly. Future use of this method in real-world applications must be coupled with appropriate clinical validation, governance, and safeguards to ensure equitable, transparent, and accountable outcomes.

\end{document}